\documentclass[11pt,a4paper]{article}
\usepackage[hyperref]{acl2018}
\usepackage{times}
\usepackage{latexsym}

\usepackage{amsmath}
\usepackage{graphicx}
\usepackage{subfigure}
\usepackage{array}
\usepackage{booktabs}
\usepackage{enumerate}
\usepackage{color}
\usepackage{url}
\usepackage{multirow}
\usepackage{xspace}
\usepackage{cite}
\usepackage{relsize}

\usepackage{amsfonts}
\usepackage{amssymb}

\newcommand{\paratitle}[1]{\noindent \textbf{#1}}
\newcommand{\tabincell}[2]{\begin{tabular}{@{}#1@{}}#2\end{tabular}}
\newcommand{\ie}{\emph{i.e.,}\xspace}
\newcommand{\eg}{\emph{e.g.,}\xspace}

\newcommand\email{\begingroup \urlstyle{tt}\smaller[2]\Url}

\aclfinalcopy

\title{\texttt{hyperdoc2vec}: Distributed Representations of Hypertext Documents}
\author{Jialong Han$^{\spadesuit}$, Yan Song$^{\spadesuit}$, Wayne Xin Zhao$^{\blacklozenge}$, Shuming Shi$^{\spadesuit}$, Haisong Zhang$^{\spadesuit}$\\
    $^\spadesuit$Tencent AI Lab\\
    $^\blacklozenge$School of Information, Renmin University of China\\
    \email{{jialonghan, batmanfly}@gmail.com,{clksong, shumingshi, hansonzhang}@tencent.com}}

\begin{document}

\maketitle

\begin{abstract}
Hypertext documents, such as web pages and academic papers, are of great importance in delivering information in our daily life.
Although being effective on plain documents, conventional text embedding methods suffer from information loss if directly adapted to hyper-documents.
In this paper, we propose a general embedding approach for hyper-documents, namely, \texttt{hyperdoc2vec},
along with four criteria characterizing necessary information that hyper-document embedding models should preserve.
Systematic comparisons are conducted between \texttt{hyperdoc2vec} and several competitors on two tasks, \ie paper classification and citation recommendation, in the academic paper domain.
Analyses and experiments both validate the superiority of \texttt{hyperdoc2vec} to other models w.r.t.\ the four criteria.
\end{abstract}

\section{Introduction}

The ubiquitous World Wide Web has boosted research interests on hypertext documents, \eg personal webpages~\citep{lu2003link}, Wikipedia pages~\citep{gabrilovich2007computing}, as well as academic papers~\citep{sugiyama2010scholarly}.
Unlike independent plain documents, a hypertext document (hyper-doc for short) links to another hyper-doc by a hyperlink or citation mark in its textual content.
Given this essential distinction, hyperlinks or citations are worth specific modeling in many tasks such as link-based classification~\citep{lu2003link}, web retrieval~\citep{page1999pagerank}, entity linking~\citep{cucerzan2007large}, and citation recommendation~\citep{he2010context}.

To model hypertext documents, various efforts~\citep{cohn2001missing,kataria2010utilizing,perozzi2014deepwalk,zwicklbauer2016robust,wang2016linked} have been made to depict networks of hyper-docs as well as their content.
Among potential techniques,
distributed representation~\citep{mikolov2013distributed,le2014distributed} tends to be promising since its validity and effectiveness are proven for plain documents on many natural language processing (NLP) tasks.

Conventional attempts on utilizing embedding techniques in hyper-doc-related tasks generally fall into two types.
The first type \citep{berger2017cite2vec,zwicklbauer2016robust} simply downcasts hyper-docs to plain documents and feeds them into \texttt{word2vec} \citep{mikolov2013distributed} (\texttt{w2v} for short) or \texttt{doc2vec}~\citep{le2014distributed} (\texttt{d2v} for short).
These approaches involve downgrading hyperlinks and inevitably omit certain information in hyper-docs.
However, no previous work investigates the information loss, and how it affects the performance of such downcasting-based adaptations.
The second type designs sophisticated embedding models to fulfill certain tasks, \eg citation recommendation~\citep{huang2015neural}, paper classification~\citep{wang2016linked}, and entity linking~\citep{yamada2016joint}, \emph{etc}.
These models are limited to specific tasks,
and it is yet unknown whether embeddings learned for those particular tasks can generalize to others.
Based on the above facts, we are interested in two questions:
\begin{itemize}
\item What information should hyper-doc embedding models preserve, and what nice property should they possess?
\item Is there a general approach to learning task-independent embeddings of hyper-docs?
\end{itemize}
%

To answer the two questions, we formalize the hyper-doc embedding task, and propose four criteria, \ie \emph{content awareness}, \emph{context awareness}, \emph{newcomer friendliness}, and \emph{context intent awareness}, to assess different models.
Then we discuss simple downcasting-based adaptations of existing approaches w.r.t.\ the above criteria, and demonstrate that none of them satisfy all four.
To this end, we propose \texttt{hyperdoc2vec} (\texttt{h-d2v} for short), a general embedding approach for hyper-docs.
Different from most existing approaches, \texttt{h-d2v} learns two vectors for each hyper-doc to characterize its roles of citing others and being cited.
Owning to this, \texttt{h-d2v} is able to directly model hyperlinks or citations without downgrading them.
To evaluate the learned embeddings, we employ two tasks in the academic paper domain\footnote{Although limited in tasks and domains, we expect that our embedding approach can be potentially generalized to, or serve as basis to more sophisticated methods for, similar tasks in the entity domain, \eg Wikipedia page classification and entity linking. We leave them for future work.}, \ie paper classification and citation recommendation.
Experimental results demonstrate the superiority of \texttt{h-d2v}.
Comparative studies and controlled experiments also confirm that \texttt{h-d2v} benefits from satisfying the above four criteria.

We summarize our contributions as follows:
\begin{itemize}
\item We propose four criteria to assess different hyper-document embedding models.
\item We propose \texttt{hyperdoc2vec}, a general embedding approach for hyper-documents.
\item We systematically conduct comparisons with competing approaches, validating the superiority of \texttt{h-d2v} in terms of the four criteria.
\end{itemize}

\section{Related Work}\label{sec:related_work}

\paratitle{Network representation learning} is a related topic to ours
since a collection of hyper-docs resemble a network.
To embed nodes in a network,
\citet{perozzi2014deepwalk} propose DeepWalk, where nodes and random walks are treated as pseudo words and texts, and fed to \texttt{w2v} for node vectors.
\citet{tang2015line} explicitly embed second-order proximity via the number of common neighbors of nodes.
\citet{grover2016node2vec} extend DeepWalk with second-order Markovian walks.
To improve classification tasks, \citet{tu2016mmdw} explore a semi-supervised setting that accesses partial labels.
Compared with these models, \texttt{h-d2v} learns from both documents' connections and contents while they mainly focus on network structures.
\vskip 0.5em

\paratitle{Document embedding for classification} is another focused area to apply document embeddings.
\citet{le2014distributed} employ learned \texttt{d2v} vectors to build different text classifiers.
\citet{tang2015pte} apply the method in~\citep{tang2015line} on word co-occurrence graphs for word embeddings, and average them for document vectors.
For hyper-docs, \citet{ganguly2017paper2vec} and \citet{wang2016linked} target paper classification in unsupervised and semi-supervised settings, respectively.
However, unlike \texttt{h-d2v}, they do not explicitly model citation contexts.
\citet{yang2015network}'s approach also addresses embedding hyper-docs, but involves matrix factorization and does not scale.
\vskip 0.5em

\paratitle{Citation recommendation} is a direct downstream task to evaluate embeddings learned for a certain kind of hyper-docs, \ie academic papers.
In this paper we concentrate on \text{context-aware} citation recommendation~\citep{he2010context}.
Some previous studies adopt neural models for this task.
\citet{huang2015neural} propose Neural Probabilistic Model (NPM) to tackle this problem with embeddings.
Their model outperforms non-embedding ones~\citep{kataria2010utilizing,tang2009discriminative,huang2012recommending}.
\citet{ebesu2017neural} also exploit neural networks for citation recommendation,
but require author information as additional input.
Compared with \texttt{h-d2v}, these models are limited in a task-specific setting.
\vskip 0.5em

\begin{figure*}[t]
  \centering
  \subfigure[Hyper-documents.]{
    \label{fig:org_view}
    \includegraphics[width=0.49\columnwidth]{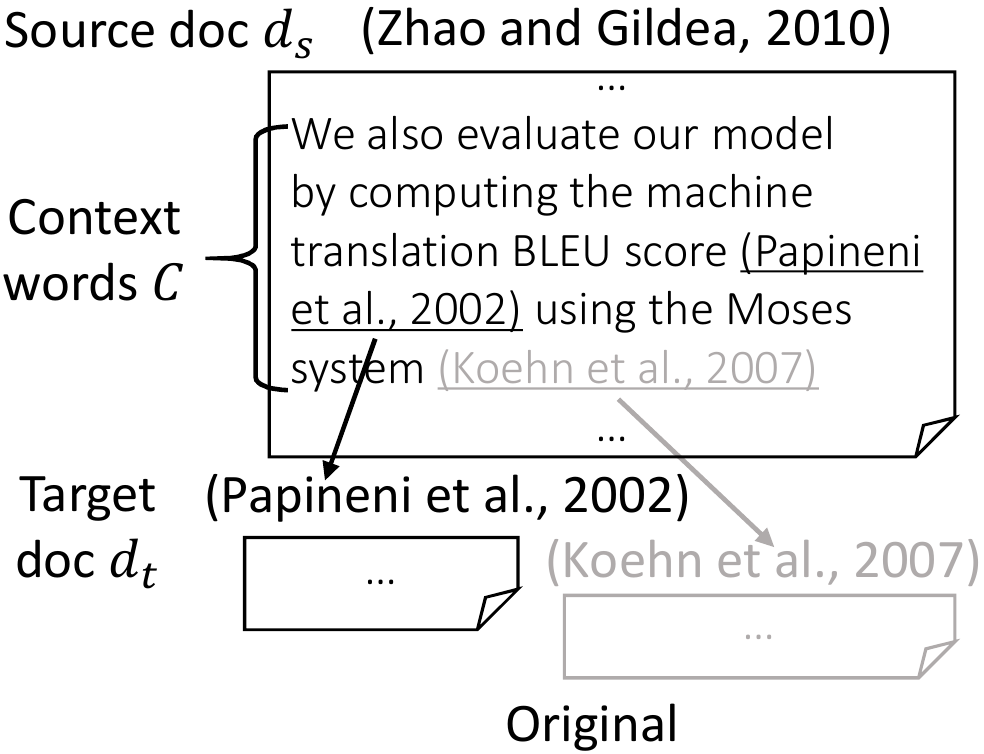}}
  \quad
  \subfigure[Citation as word.]{
    \label{fig:caw_view}
    \includegraphics[width=0.71\columnwidth]{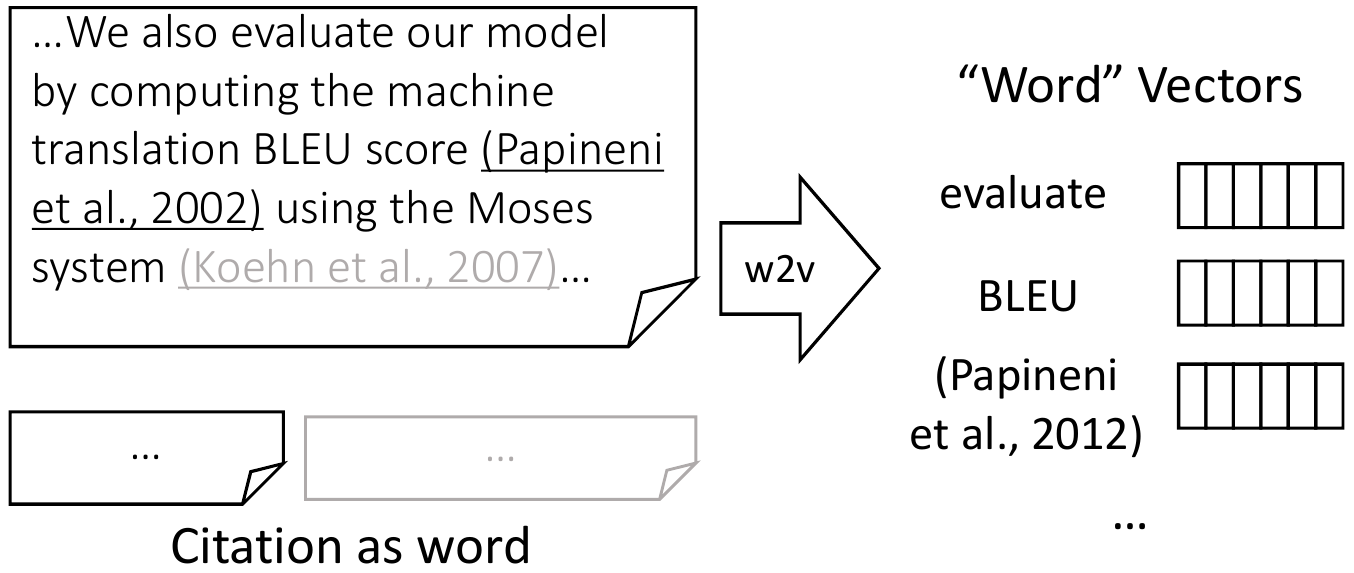}}
  \quad
  \subfigure[Context as content.]{
    \label{fig:cac_view}
    \includegraphics[width=0.71\columnwidth]{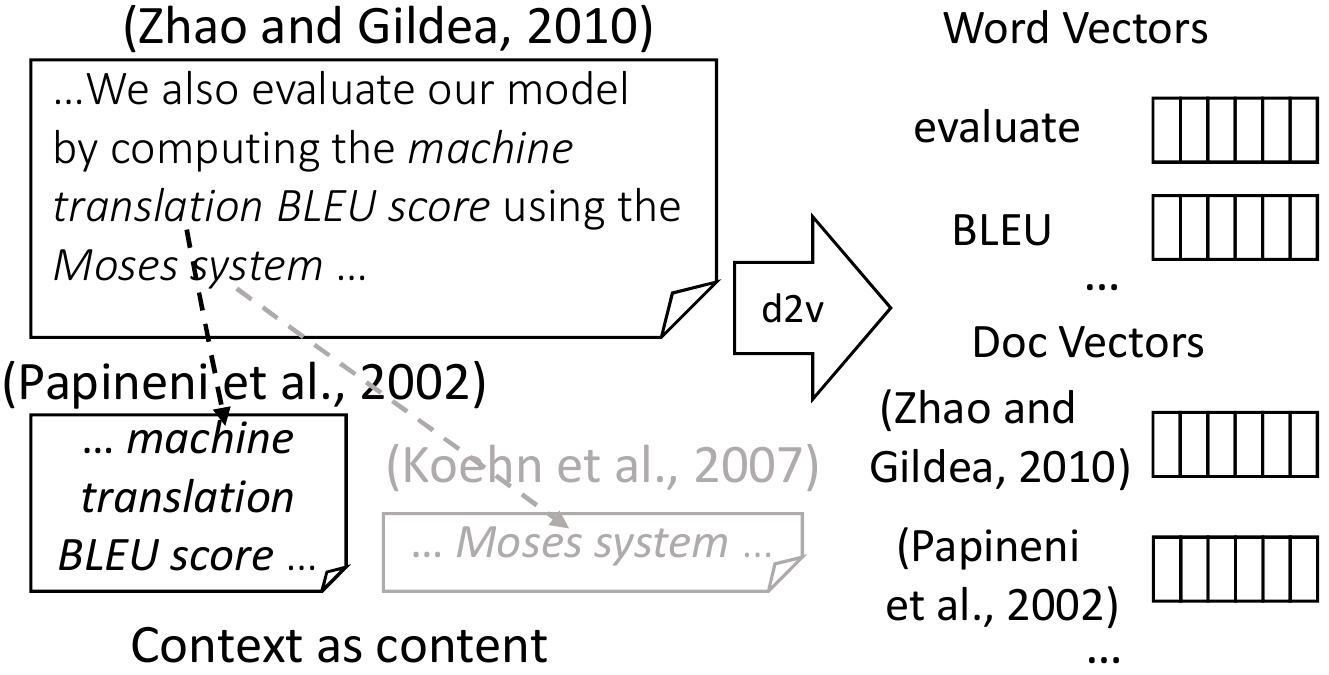}}
  \caption{An example of~\citet{zhao2010fast} citing~\citet{papineni2002bleu} and existing approaches.}
\end{figure*}

\paratitle{Embedding-based entity linking}
%
is another topic that exploits embeddings to model certain hyper-docs, \ie
Wikipedia \citep{huang2015leveraging,yamada2016joint,sun2015modeling,fang2016entity,he2013learning,zwicklbauer2016robust}, for entity linking~\citep{shen2015entity}.
It resembles citation recommendation in the sense that linked entities highly depend on the contexts.
Meanwhile, it requires extra steps like candidate generation, and can benefit from sophisticated techniques such as collective linking \citep{cucerzan2007large}.


\section{Preliminaries}

We introduce notations and definitions, then formally define the embedding problem.
We also propose four criteria for hyper-doc embedding models w.r.t\ their appropriateness and informativeness.


\subsection{Notations and Definitions}

Let $w\in W$ be a word from a vocabulary $W$, and $d\in D$ be a \emph{document id} (\eg web page URLs and paper DOIs) from an id collection $D$.
After filtering out non-textual content,
a \emph{hyper-document} $H$ is reorganized as a sequence of words and doc ids, \ie $W\cup D$.
For example, web pages could be simplified as streams of words and URLs, and papers are actually sequences of words and cited DOIs.

If a document id $d_t$ with some surrounding words $C$ appear in the hyper-doc of $d_s$, \ie $H_{d_s}$, we stipulate that a \emph{hyper-link} $\langle d_s,C,d_t\rangle$ is formed.
Herein $d_s,d_t\in D$ are ids of the \emph{source} and \emph{target} documents, respectively; $C\subseteq W$ are \emph{context words}.
Figure~\ref{fig:org_view} exemplifies a hyperlink.

\subsection{Problem Statement}

Given a corpus of hyper-docs $\{H_d\}_{d\in D}$ with $D$ and $W$, we want to learn document and word embedding matrices $\mathbf{D}\in\mathbb{R}^{k\times|D|}$ and $\mathbf{W}\in\mathbb{R}^{k\times|W|}$ simultaneously.
The $i$-th column $\mathbf{d}_i$ of $\mathbf{D}$ is a $k$-dimensional embedding vector for the $i$-th hyper-doc with id $d_i$.
Similarly, $\mathbf{w}_j$, the $j$-th column of $\mathbf{W}$, is the vector for word $w_j$.
Once embeddings for hyper-docs and words are learned, they can facilitate applications like hyper-doc classification and citation recommendation.

\subsection{Criteria for Embedding Models}\label{ssec:property}

A reasonable model should learn how contents and hyperlinks in hyper-docs impact both $\mathbf{D}$ and $\mathbf{W}$.
We propose the following criteria for models:
\begin{itemize}
\item \textbf{Content aware.}
Content words of a hyper-doc play the main role in describing it, so the document representation should depend on its own content.
For example, the words in \citet{zhao2010fast} should affect and contribute to its embedding.
\item \textbf{Context aware.} Hyperlink contexts usually provide a summary for the target document.
Therefore, the target document's vector should be impacted by words that others use to summarize it, \eg paper \citet{papineni2002bleu} and the word ``\emph{BLEU}'' in Figure~\ref{fig:org_view}.
\item \textbf{Newcomer friendly.}
In a hyper-document network, it is inevitable that some documents are not referred to by any hyperlink in other hyper-docs.
If such ``newcomers'' do not get embedded properly, downstream tasks involving them are infeasible or deteriorated.
\item \textbf{Context intent aware.}
Words around a hyperlink, \eg ``evaluate \dots by'' in Figure~\ref{fig:org_view}, normally indicate why the source hyper-doc makes the reference, \eg for general reference or to follow/oppose the target hyper-doc's opinion or practice.
Vectors of those context words should be influenced by both documents to characterize such semantics or intents between the two documents.
\end{itemize}
We note that the first three criteria are for hyper-docs, while the last one is desired for word vectors.

\section{Representing Hypertext Documents}
\label{sec:embedding}

In this section, we first give the background of two prevailing techniques, \texttt{word2vec} and \texttt{doc2vec}.
Then we present two conversion approaches for hyper-documents so that \texttt{w2v} and \texttt{d2v} can be applied.
Finally, we address their weaknesses w.r.t. the aforementioned four criteria, and propose our \texttt{hyperdoc2vec} model.
In the remainder of this paper, when the context is clear, we mix the use of terms hyper-doc/hyperlink with paper/citation.

\subsection{\texttt{word2vec} and \texttt{doc2vec}}

\begin{table*}\small
\setlength\tabcolsep{2pt}
\renewcommand{\arraystretch}{1.16}
\parbox{.67\linewidth}{
  \centering
  \begin{tabular}{m{2.7cm}|>{\centering\arraybackslash}m{2cm}>{\centering\arraybackslash}m{2cm}|cccc}
    \toprule
    \multirow{3}{*}{\textbf{Desired Property}}&\multicolumn{2}{c|}{\textbf{Impacts Task?}}&\multicolumn{4}{c}{\textbf{Addressed by Approach?}}\\
    \cline{2-7}
    &Classification&Citation Recommendation&w2v&d2v-nc&d2v-cac&h-d2v\\
    \midrule
    Context aware&\checkmark&\checkmark&\checkmark&$\times$&\checkmark&\checkmark\\
    Content aware&\checkmark&\checkmark&$\times$&\checkmark&\checkmark&\checkmark\\
    Newcomer friendly&\checkmark&\checkmark&$\times$&\checkmark&\checkmark&\checkmark\\
    Context intent aware&$\times$&\checkmark&$\times$&$\times$&$\times$&\checkmark\\
    \bottomrule
  \end{tabular}
  \caption{Analysis of tasks and approaches w.r.t.\ desired properties.}\label{tab:comparison}
  }
  \quad
  \parbox{.27\linewidth}{
  \setlength\tabcolsep{2.5pt}
  \renewcommand{\arraystretch}{1.25}
  \centering
  \begin{tabular}{l|cccc}
  \toprule
  \multirow{2}{*}{\textbf{Model}}&\multicolumn{4}{c}{\textbf{Output}}\\
  \cmidrule{2-5}
  &$\mathbf{D}^I$&$\mathbf{D}^O$&$\mathbf{W}^I$&$\mathbf{W}^O$\\
  \midrule
  w2v&\checkmark&\checkmark&\checkmark&\checkmark\\
  d2v (pv-dm)&\checkmark&$\times$&\checkmark&\checkmark\\
  d2v (pv-dbow)&\checkmark&$\times$&$\times$&\checkmark\\
  h-d2v&\checkmark&\checkmark&\checkmark&\checkmark\\
  \bottomrule
  \end{tabular}
  \caption{Output of models.}\label{tab:output}
  }
\end{table*}

\texttt{w2v}~\citep{mikolov2013distributed} has proven effective for many NLP tasks.
It integrates two models, \ie \texttt{cbow} and \texttt{skip-gram}, both of which learn two types of word vectors, \ie IN and OUT vectors.
\texttt{cbow} sums up IN vectors of context words and make it predictive of the current word's OUT vector.
\texttt{skip-gram} uses the IN vector of the current word to predict its context words' OUT vectors.

As a straightforward extension to \texttt{w2v}, \texttt{d2v} also has two variants: \texttt{pv-dm} and \texttt{pv-dbow}.
\texttt{pv-dm} works in a similar manner as \texttt{cbow}, except that the IN vector of the current document is regarded as a special context vector to average.
Analogously, \texttt{pv-dbow} uses IN document vector to predict its words' OUT vectors, following the same structure of \texttt{skip-gram}.
Therefore in \texttt{pv-dbow}, words' IN vectors are omitted.

\subsection{Adaptation of Existing Approaches}\label{ssec:existing_approaches}

To represent hyper-docs, a straightforward strategy is to convert them into plain documents in a certain way and apply \texttt{w2v} and \texttt{d2v}.
Two conversions following this strategy are illustrated below.
\vskip 0.5em

\paratitle{Citation as word.} This approach is adopted by \citet{berger2017cite2vec}.\footnote{It is designed for document visualization purposes.}
As Figure~\ref{fig:caw_view} shows, document ids $D$ are treated as a collection of special words.
Each citation is regarded as an occurrence of the target document's special word.
After applying standard word embedding methods, \eg\texttt{w2v}, we obtain embeddings for both ordinary words and special ``words'', \ie documents.
In doing so, this approach allows target documents interacting with context words, thus produces context-aware embeddings for them.
\vskip 0.5em

\paratitle{Context as content.} It is often observed in academic papers when citing others' work, an author briefly summarizes the cited paper in its citation context.
Inspired by this, we propose a context-as-content approach as in Figure~\ref{fig:cac_view}.
To start, we remove all citations.
Then all citation contexts of a target document $d_t$ are copied into $d_t$ as additional contents to make up for the lost information.
Finally, \texttt{d2v} is applied to the augmented documents to generate document embeddings.
%
With this approach, the generated document embeddings are both context- and content-aware.

\subsection{\texttt{hyperdoc2vec}}

Besides citation-as-word with \texttt{w2v} and context-as-content with \texttt{d2v} (denoted by \texttt{d2v-cac} for short),
there is also an alternative using \texttt{d2v} on documents with citations removed (\texttt{d2v-nc} for short).
We made a comparison of these approaches in Table \ref{tab:comparison} in terms of the four criteria stated in Section~\ref{ssec:property}.
%
It is observed that none of them satisfy all criteria, where the reasons are as follows.

First, \texttt{w2v} is not content aware.
Following our examples in the academic paper domain,
consider the paper (hyper-doc) \citet{zhao2010fast} in Figure~\ref{fig:org_view},
from \texttt{w2v}'s perspective in Figure~\ref{fig:caw_view}, ``\dots computing the machine translation BLEU \dots'' and other text no longer have association with \citet{zhao2010fast}, thus not contributing to its embedding.
In addition, for papers being just published and having not obtained citations yet, they will not appear as special ``words'' in any text.
This makes \texttt{w2v} newcomer-unfriendly, \ie unable to produce embeddings for them.
Second, being trained on a corpus without citations, \texttt{d2v-nc} is obviously not context aware.
Finally, in both \texttt{w2v} and \texttt{d2v-cac}, context words interact with the target documents without treating the source documents as backgrounds,
which forces IN vectors of words with context intents, \eg ``\textit{evaluate}'' and ``\textit{by}'' in Figure~\ref{fig:org_view}, to simply remember the target documents, rather than capture the semantics of the citations.

The above limitations are caused by the conversions of hyper-docs where certain information in citations is lost.
For a citation $\langle d_s,C,d_t\rangle$, citation-as-word only keeps the co-occurrence information between $C$ and $d_t$.
Context-as-content, on the other hand, mixes $C$ with the original content of $d_t$.
Both approaches implicitly downgrade citations $\langle d_s,C,d_t\rangle$ to $\langle C,d_t\rangle$ for adaptation purposes.

%
To learn hyper-doc embeddings without such limitations, we propose \texttt{hyperdoc2vec}.
In this model, two vectors of a hyper-doc $d$, \ie IN and OUT vectors, are adopted to represent the document of its two roles.
The IN vector $\mathbf{d}^I$ characterizes $d$ being a source document.
The OUT vector $\mathbf{d}^O$ encodes its role as a target document.
We note that learning those two types of vectors is advantageous.
It enables us to model citations and contents simultaneously without sacrificing information on either side.
Next, we describe the details of \texttt{h-d2v} in modeling citations and contents.

\begin{figure}
  \centering
  \includegraphics[width=\columnwidth]{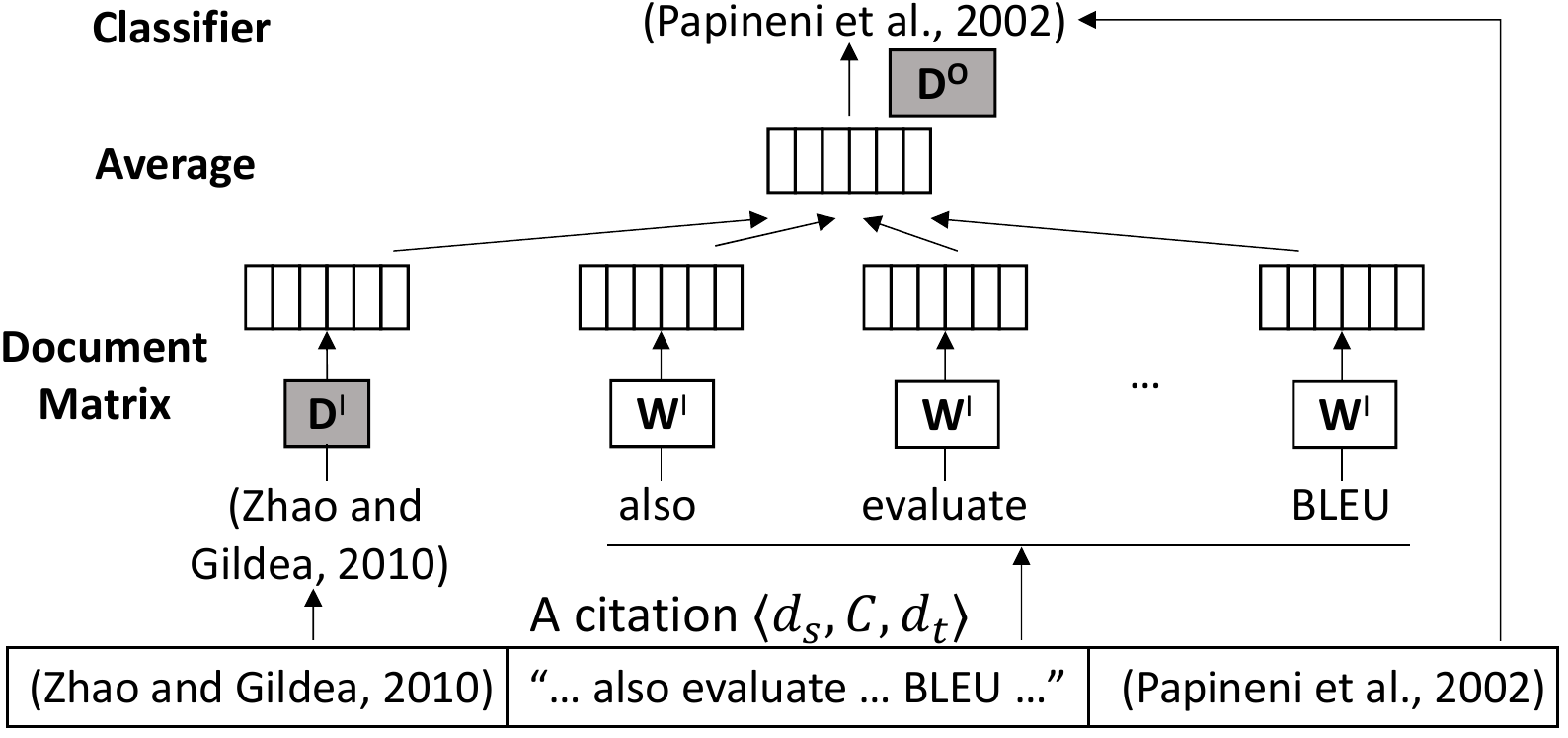}
  \caption{The \texttt{hyperdoc2vec} model.}\label{fig:model}
\end{figure}


To model citations, we adopt the architecture in Figure \ref{fig:model}.
It is similar to \texttt{pv-dm}, except that documents rather than words are predicted at the output layer.
For a citation $\langle d_s,C,d_t\rangle$, to allow context words $C$ interacting with both vectors, we average $\mathbf{d}^I_s$ of $d_s$ with word vectors of $C$, and make the resulted vector predictive of $\mathbf{d}^O_t$ of $d_t$.
Formally, for all citations $\mathcal{C}=\{\langle d_s,C,d_t\rangle\}$, we aim to optimize the following average log probability objective:
\begin{equation}\label{eq:obj}
\max_{\mathbf{D}^I,\mathbf{D}^O,\mathbf{W}^I}\quad\frac{1}{|\mathcal{C}|}\sum_{\langle d_s,C,d_t\rangle\in \mathcal{C}}\log P(d_t|d_s,C)
\end{equation}
To model the probability $P(d_t|d_s,C)$ where $d_t$ is cited in $d_s$ with $C$, we average their IN vectors
\begin{equation}
\textbf{x}=\frac{1}{1+|C|}\biggl(\textbf{d}_s^I+\sum_{w\in C}\textbf{w}^I\biggr)\label{eq:average}
\end{equation}
and use $\mathbf{x}$ to compose a multi-class softmax classifier on all OUT document vectors
\begin{equation}
P(d_t|d_s,C)=\frac{\exp(\textbf{x}^\top\textbf{d}_t^O)}{\sum_{d\in D}\exp(\textbf{x}^\top\textbf{d}^O)}\label{eq:soft_max}\\
\end{equation}


%

To model contents' impact on document vectors, we simply consider an additional objective function that is identical to \texttt{pv-dm}, \ie enumerate words and contexts, and use the same input architecture as Figure~\ref{fig:model} to predict the OUT vector of the current word.
Such convenience owes to the fact that using two vectors makes the model parameters compatible with those of \texttt{pv-dm}.
Note that combining the citation and content objectives leads to a joint learning framework.
To facilitate easier and faster training, we adopt an alternative pre-training/fine-tuning or \emph{retrofitting} framework~\citep{faruqui2015retrofitting}.
We initialize with a predefined number of \texttt{pv-dm} iterations, and then optimize Eq.~\ref{eq:obj} based on the initialization.

Finally, similar to \texttt{w2v}~\citep{mikolov2013distributed} and \texttt{d2v}~\citep{le2014distributed}, to make training efficient, we adopt negative sampling:
\begin{equation}\label{eq:negative_sampling}
\log\sigma(\textbf{x}^\top\textbf{d}_t^O)+\sum_{i=1}^{n}\mathbb{E}_{d_i\sim P_N(d)}\log\sigma(-\textbf{x}^\top\textbf{d}_i^O)
\end{equation}
and use it to replace every $\log P(d_t|d_s,C)$.
Following \citet{huang2015neural}, we adopt a uniform distribution on $D$ as the distribution $P_N(d)$.

Unlike the other models in Table~\ref{tab:comparison}, \texttt{h-d2v} satisfies all four criteria.
We refer to the example in Figure~\ref{fig:model} to make the points clear.
First, when optimizing Eq.~\ref{eq:obj} with the instance in Figure~\ref{fig:model}, the update to $\mathbf{d}^O$ of \citet{papineni2002bleu} depends on $\mathbf{w}^I$ of context words such as ``\textit{BLEU}''.
Second, we pre-train $\mathbf{d}^I$ with contents, which makes the document embeddings content aware.
Third, newcomers can depend on their contents for $\mathbf{d}^I$, and update their OUT vectors when they are sampled\footnote{Given a relatively large $n$.} in Eq.~\ref{eq:negative_sampling}.
Finally, the optimization of Eq.~\ref{eq:obj} enables mutual enhancement between vectors of hyper-docs and context intent words, \eg ``\textit{evaluate by}''.
Under the background of a machine translation paper~\citet{zhao2010fast}, the above two words help point the citation to the BLEU paper~\citep{papineni2002bleu}, thus updating its OUT vector.
The intent ``\emph{adopting tools/algorithms}'' of ``evaluate by'' is also better captured by iterating over many document pairs with them in between.

%


\begin{table}\small
  \centering
  \begin{tabular}{l|l|rrr}
    \toprule
    \multicolumn{2}{c|}{\textbf{Dataset}}&\textbf{Docs}&\textbf{Citations}&\textbf{Years}\\
    \midrule
    \multirow{3}{*}{NIPS}&Train&1,590&512&Up to 1998\\
    &Test&150&89&1999\\
    &Total&1,740&601&Up to 1999\\
    \midrule
    \multirow{3}{*}{ACL}&Train&18,845&91,792&Up to 2012\\
    &Test&1,563&16,937&2013\\
    &Total&20,408&108,729&Up to 2013\\
    \midrule
    \multirow{3}{*}{DBLP}&Train&593,378&2,565,625&Up to 2009\\
    &Test&55,736&308,678&From 2010\\
    &Total&649,114&2,874,303&All years\\
    \bottomrule
  \end{tabular}
  \caption{The statistics of three datasets.}\label{tab:data}
\end{table}

\section{Experiments}

\begin{table*}\small
\setlength\tabcolsep{4.2pt}
\parbox{.4\linewidth}{
\renewcommand{\arraystretch}{1.06}
  \centering
  \begin{tabular}{l|cc|cc}
    \toprule
    \multirow{2}{*}{\textbf{Model}}&\multicolumn{2}{c|}{\textbf{Original}}&\multicolumn{2}{c}{\textbf{w/ DeepWalk}}\\
    \cline{2-5}
    &Macro&Micro&Macro&Micro\\
    \midrule
    DeepWalk&61.67&69.89&61.67&69.89\\
    w2v (I)&10.83&41.84&31.06&50.93\\
    w2v (I+O)&9.36&41.26&25.92&49.56\\
    d2v-nc&70.62&77.86&70.64&78.06\\
    d2v-cac&71.83&78.09&71.57&78.59\\
    \midrule
    h-d2v (I)&68.81&76.33&\textbf{73.96}&\textbf{79.93}\\
    h-d2v (I+O)&\textbf{72.89}&\textbf{78.99}&73.24&79.55\\
    \bottomrule
  \end{tabular}
  \caption{F\textsubscript{1} scores on DBLP.\label{tab:classification}}
  }
  \quad
\parbox{.55\linewidth}{
  \centering
  \begin{tabular}{l|c|cc|cc}
    \toprule
    \multirow{2}{*}{\textbf{Model}}&\textbf{Content Aware/}&\multicolumn{2}{c|}{\textbf{Original}}&\multicolumn{2}{c}{\textbf{w/ DeepWalk}}\\
    \cline{3-6}
    &\textbf{Newcomer Friendly}&Macro&Micro&Macro&Micro\\
    \midrule
    DeepWalk&-&66.57&\textbf{76.56}&66.57&76.56\\
    \midrule
    w2v (I)&$\times$ / $\times$&19.77&47.32&59.80&72.90\\
    w2v (I+O)&$\times$ / $\times$&15.97&45.66&50.77&70.08\\
    \midrule
    d2v-nc&\checkmark / \checkmark&61.54&73.73&69.37&78.22\\
    d2v-cac&\checkmark / \checkmark&65.23&75.93&\textbf{70.43}&\textbf{78.75}\\
    h-d2v (I)&\checkmark / \checkmark&58.59&69.79&66.99&75.63\\
    h-d2v (I+O)&\checkmark / \checkmark&\textbf{66.64}&75.19&68.96&76.61\\
    \bottomrule
  \end{tabular}
  \caption{F\textsubscript{1} on DBLP when newcomers are discarded.\label{tab:cited_doc_only_clf}}
}
\end{table*}

In this section, we first introduce datasets and basic settings used to learn embeddings.
We then discuss additional settings and present experimental results of the two tasks, \ie document classification and citation recommendation, respectively.

\subsection{Datasets and Experimental Settings}

We use three datasets from the academic paper domain, \ie NIPS\footnote{https://cs.nyu.edu/~roweis/data.html}, ACL anthology\footnote{http://clair.eecs.umich.edu/aan/index.php ~(2013 release)} and DBLP\footnote{http://zhou142.myweb.cs.uwindsor.ca/academicpaper.html This page has been unavailable recently. They provide a larger CiteSeer dataset and a collection of DBLP paper ids. To better interpret results from the Computer Science perspective, we intersect them and obtain the DBLP dataset.}, as shown in Table~\ref{tab:data}.
They all contain full text of papers, and are of small, medium, and large size, respectively.
We apply ParsCit\footnote{https://github.com/knmnyn/ParsCit}~\citep{councill2008parscit} to parse the citations and bibliography sections.
Each identified citation string referring to a paper in the same dataset, \eg [1] or (Author et al., 2018), is replaced by a global paper id.
Consecutive citations like [1, 2] are regarded as multiple ground truths occupying one position.
%
Following \citet{he2010context}, we take 50 words before and after a citation as the citation context.

Gensim~\citep{rehurek_lrec} is used to implement all \texttt{w2v} and \texttt{d2v} baselines as well as \texttt{h-d2v}.
We use \texttt{cbow} for \texttt{w2v} and \texttt{pv-dbow} for \texttt{d2v}, unless otherwise noted.
For all three baselines, we set the (half) context window length to 50.
For \texttt{w2v}, \texttt{d2v}, and the \texttt{pv-dm}-based initialization of \texttt{h-d2v}, we run 5 epochs following Gensim's default setting.
For \texttt{h-d2v}, its iteration is set to 100 epochs with 1000 negative samples.
The dimension size $k$ of all approaches is 100.
All other parameters in Gensim are kept as default.

\subsection{Document Classification}

In this task, we classify the research fields of papers given their vectors learned on DBLP.
To obtain labels, we use Cora\footnote{http://people.cs.umass.edu/\textasciitilde mccallum/data.html}, a small dataset of Computer Science papers and their field categories.
We keep the first levels of the original categories, \eg ``Artificial Intelligence'' of ``Artificial Intelligence - Natural Language Processing'', leading to 10 unique classes.
We then intersect the dataset with DBLP, and obtain 5,975 labeled papers.

For \texttt{w2v} and \texttt{h-d2v} outputing both IN and OUT document vectors, we use IN vectors or concatenations of both vectors as features.
For newcomer papers without \texttt{w2v} vectors, we use zero vectors instead.
To enrich the features with network structure information, we also try concatenating them with the output of DeepWalk~\citep{perozzi2014deepwalk}, a representative network embedding model.
The model is trained on the citation network of DBLP with an existing implementation\footnote{https://github.com/phanein/deepwalk} and default parameters.
An SVM classifier with RBF kernel is used.
We perform 5-fold cross validation, and report Macro- and Micro-F\textsubscript{1} scores.

\subsubsection{Classification Performance}\label{ssec:clf_perf}

\begin{table*}\small
  \centering
  \setlength\tabcolsep{3.3pt}
  \begin{tabular}{l|cccc|cccc|cccc}
    \toprule
    \multirow{2}{*}{\textbf{Model}}&\multicolumn{4}{c|}{\textbf{NIPS}}&\multicolumn{4}{c|}{\textbf{ACL Anthology}}&\multicolumn{4}{c}{\textbf{DBLP}}\\
    \cline{2-13}
    &Rec&MAP&MRR&nDCG&Rec&MAP&MRR&nDCG&Rec&MAP&MRR&nDCG\\
    \midrule
    w2v (cbow, I4I)&5.06&1.29&1.29&2.07&12.28&5.35&5.35&6.96&3.01&1.00&1.00&1.44\\
    w2v (cbow, I4O)&12.92&\textbf{6.97}&\textbf{6.97}&8.34&15.68&8.54&8.55&10.23&13.26&7.29&7.33&8.58\\
    d2v-nc (pv-dbow, cosine)&14.04&3.39&3.39&5.82&21.09&9.65&9.67&12.29&7.66&3.25&3.25&4.23\\
    d2v-cac (same as d2v-nc)&14.61&4.94&4.94&7.14&28.01&11.82&11.84&15.59&15.67&7.34&7.36&9.16\\
    NPM~\citep{huang2015neural}&7.87&2.73&3.13&4.03&12.86&5.98&5.98&7.59&6.87&3.28&3.28&4.07\\
    \midrule
    h-d2v (random init, I4O)&3.93&0.78&0.78&1.49&30.98&16.76&16.77&20.12&17.22&8.82&8.87&10.65\\
    h-d2v (pv-dm retrofitting, I4O)&\textbf{15.73}&6.68&6.68&\textbf{8.80}&\textbf{31.93}&\textbf{17.33}&\textbf{17.34}&\textbf{20.76}&\textbf{21.32}&\textbf{10.83}&\textbf{10.88}&\textbf{13.14}\\
    \bottomrule
  \end{tabular}
  \caption{Top-10 citation recommendation results (dimension size $k=100$).}\label{tab:citation_recom}
\end{table*}

In Table~\ref{tab:classification}, we demonstrate the classification results.
We have the following observations.

First, adding DeepWalk information almost always leads to better classification performance, except for Macro-F\textsubscript{1} of the \texttt{d2v-cac} approach.

Second, owning to different context awareness, \texttt{d2v-cac} consistently outperforms \texttt{d2v-nc} in terms of all metrics and settings.

Third, \texttt{w2v} has the worst performance.
The reason may be that \texttt{w2v} is neither content aware nor newcomer friendly.
We will elaborate more on the impacts of the two properties in Section~\ref{ssec:ctrl_exp_clf}.

Finally, no matter whether DeepWalk vectors are used, \texttt{h-d2v} achieves the best F\textsubscript{1} scores.
However, when OUT vectors are involved, \texttt{h-d2v} with DeepWalk has slightly worse performance.
A possible explanation is that, when \texttt{h-d2v} IN and DeepWalk vectors have enough information to train the SVM classifiers, adding another 100 features (OUT vectors) only increase the parameter space of the classifiers and the training variance.
For \texttt{w2v} with or without DeepWalk, it is also the case.
This may be because information in \texttt{w2v}'s IN and OUT vectors is fairly redundant.

\subsubsection{Impacts of Content Awareness and Newcomer Friendliness}\label{ssec:ctrl_exp_clf}

Because content awareness and newcomer friendliness are highly correlated in Table~\ref{tab:comparison}, to isolate and study their impacts, we decouple them as follows.
In the 5,975 labeled papers, we keep 2,052 with at least one citation, and redo experiments in Table~\ref{tab:classification}.
By carrying out such controlled experiments, we expect to remove the impact of newcomers, and compare all approaches only with respect to different content awareness.
In Table~\ref{tab:cited_doc_only_clf}, we provide the new scores obtained.

By comparing Tables~\ref{tab:classification} and~\ref{tab:cited_doc_only_clf}, we observe that \texttt{w2v} benefits from removing newcomers with zero vectors, while all newcomer friendly approaches get lower scores because of fewer training examples.
Even though the change, \texttt{w2v} still cannot outperform the other approaches, which reflects the positive impact of content awareness on the classification task.
It is also interesting that DeepWalk becomes very competitive.
This implies that structure-based methods favor networks with better connectivity.
Finally, we note that Table~\ref{tab:cited_doc_only_clf} is based on controlled experiments with intentionally skewed data.
The results are not intended for comparison among approaches in practical scenarios.

\subsection{Citation Recommendation}\label{ssec:cit_recom}

When writing papers, it is desirable to recommend proper citations for a given context.
This could be achieved by comparing the vectors of the context and previous papers.
We use all three datasets for this task.
Embeddings are trained on papers before 1998, 2012, and 2009, respectively.
The remaining papers in each dataset are used for testing.

We compare \texttt{h-d2v} with all approaches in Section~\ref{ssec:existing_approaches}, as well as NPM\footnote{Note that the authors used $n=1000$ for negative sampling, and did not report the number of training epoches.
After many trials, we find that setting the number of both the negative samples and epoches at 100 to be relatively effective and affordable w.r.t. training time.}~\citep{huang2015neural} mentioned in Section~\ref{sec:related_work}, the first embedding-based approach for the citation recommendation task.
Note that the inference stage involves interactions between word and document vectors and is non-trivial.
We describe our choices as below.

First, for \texttt{w2v} vectors, \citet{nalisnick2016improving} suggest that the IN-IN similarity favors word pairs with similar functions (\eg ``red'' and ``blue''), while the IN-OUT similarity characterizes word co-occurrence or compatibility (\eg ``red'' and ``bull'').
For citation recommendation that relies on the compatibility between context words and cited papers, we hypothesize that the IN-for-OUT (or I4O for short) approach will achieve better results.
Therefore, for \texttt{w2v}-based approaches, we average IN vectors of context words, then score and and rank OUT document vectors by dot product.

Second, for \texttt{d2v}-based approaches, we use the learned model to infer a document vector $\mathbf{d}$ for the context words, and use $\mathbf{d}$ to rank IN document vectors by cosine similarity.
Among multiple attempts, we find this choice to be optimal.

Third, for \texttt{h-d2v}, we adopt the same scoring and ranking configurations as for \texttt{w2v}.

Finally, for NPM, we adopt the same ranking strategy as in \citet{huang2015neural}.
Following them, we focus on top-10 results and report the Recall, MAP, MRR, and nDCG scores.

\subsubsection{Recommendation Performance}

\begin{figure}
  \centering
  \includegraphics[width=0.8\columnwidth]{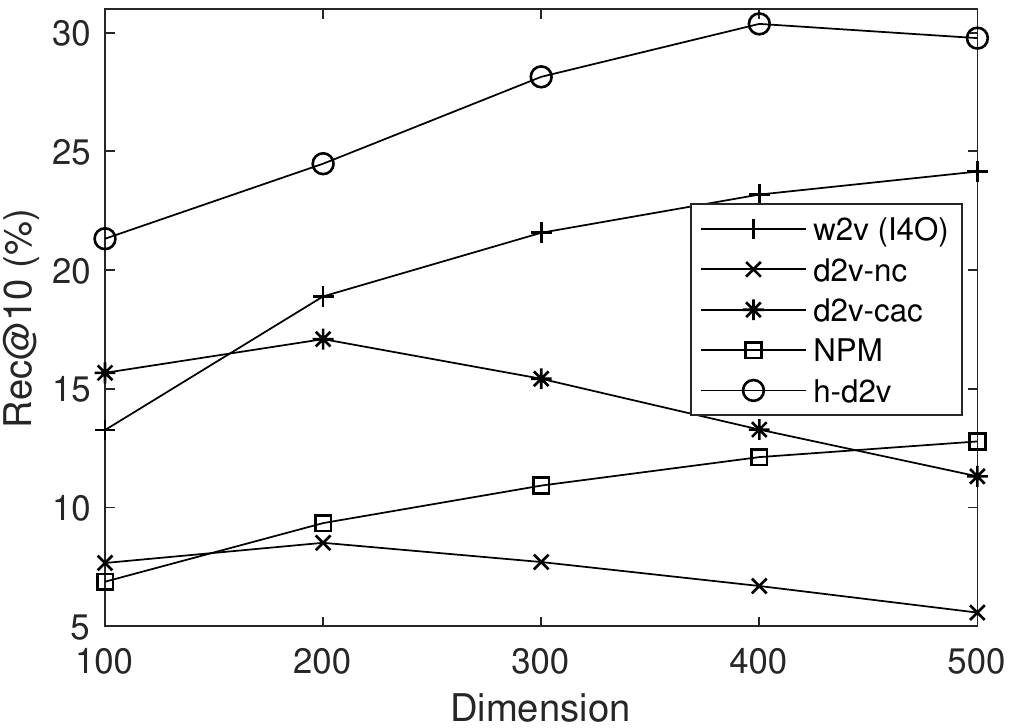}
  \caption{Varying $k$ on DBLP. The scores of \texttt{w2v} keeps increasing to 26.63 at $k=1000$, and then begins to drop. Although at the cost of a larger model and longer training/inference time, it still cannot outperform \texttt{h-d2v} of 30.37 at $k=400$.}\label{fig:dim}
\end{figure}

In Table~\ref{tab:citation_recom}, we report the citation recommendation
results. Our observations are as follows.

First, among all datasets, all methods perform relatively well on the medium-sized ACL dataset.
This is because the smallest NIPS dataset provides too few citation contexts to train a good model.
Moreover, DBLP requires a larger dimension size $k$ to store more information in the embedding vectors.
We increase $k$ and report the Rec@10 scores in Figure~\ref{fig:dim}.
We see that all approaches have better performance when $k$ increases to 200, though \texttt{d2v}-based ones start to drop beyond this point.

Second, the I4I variant of \texttt{w2v} has the worst performance among all approaches.
This observation validates our hypothesis in Section~\ref{ssec:cit_recom}.

Third, the \texttt{d2v-cac} approach outperforms its variant \texttt{d2v-nc} in terms of all datasets and metrics.
This indicates that context awareness matters in the citation recommendation task.

Fourth, the performance of NPM is sandwiched between those of \texttt{w2v}'s two variants.
We have tried our best to reproduce it.
Our explanation is that NPM is citation-as-word-based, and only depends on citation contexts for training.
Therefore, it is only context aware but neither content aware nor newcomer friendly, and behaves like \texttt{w2v}.

Finally, when retrofitting \texttt{pv-dm}, \texttt{h-d2v} generally has the best performance.
When we substitute \texttt{pv-dm} with random initialization, the performance is deteriorated by varying degrees on different datasets.
This implies that content awareness is also important, if not so important than context awareness, on the citation recommendation task.


\subsubsection{Impact of Newcomer Friendliness}

\begin{table}\small
  \setlength\tabcolsep{4pt}
  \centering
  \begin{tabular}{l|>{\centering\arraybackslash}m{1.5cm}|cccc}
    \hline
\toprule
\textbf{Model}&\textbf{Newcomer Friendly}&\textbf{Rec\textbf}&\textbf{MAP}&\textbf{MRR}&\textbf{nDCG}\\
\midrule
w2v (I4O)&$\times$&3.64&3.23&3.41&2.73\\
NPM&$\times$&1.37&1.13&1.15&0.92\\
\midrule
d2v-nc&\checkmark&6.48&3.52&3.54&3.96\\
d2v-cac&\checkmark&\textbf{8.16}&\textbf{5.13}&\textbf{5.24}&\textbf{5.21}\\
h-d2v&\checkmark&6.41&4.95&5.21&4.49\\
\bottomrule
  \end{tabular}
  \caption{DBLP results evaluated on 63,342 citation contexts with newcomer ground-truth.}\label{tab:uncited_doc_only_citation_recom}
\end{table}

Table~\ref{tab:uncited_doc_only_citation_recom} analyzes the impact of newcomer friendliness.
Opposite from what is done in Section~\ref{ssec:ctrl_exp_clf}, we only evaluate on testing examples where at least a ground-truth paper is a newcomer.
Please note that newcomer unfriendly approaches do not necessarily get zero scores.
The table shows that newcomer friendly approaches are superior to unfriendly ones.
Note that, like Table~\ref{tab:cited_doc_only_clf}, this table is also based on controlled experiments and not intended for comparing approaches.

\begin{table}\small
  \setlength\tabcolsep{2pt}
  \centering
  \begin{tabular}{lp{6.2cm}}
    \toprule
    \textbf{Category}&\textbf{Description}\\
    \midrule
    Weak&Weakness of cited approach\\
    \midrule
    CoCoGM&Contrast/Comparison in Goals/Methods (neutral)\\
    CoCo-&Work stated to be superior to cited work\\
    CoCoR0&Contrast/Comparison in Results (neutral)\\
    CoCoXY&Contrast between 2 cited methods\\
    \midrule
    PBas&Author uses cited work as basis or starting point\\
    PUse&Author uses tools/algorithms/data/definitions\\
    PModi&Author adapts or modifies tools/algorithms/data\\
    PMot&This citation is positive about approach used or problem addressed (used to motivate work in current paper)\\
    PSim&Author's work and cited work are similar\\
    PSup&Author's work and cited work are compatible/provide support for each other\\
    Neut&Neutral description of cited work, or not enough textual evidence for above categories, or unlisted citation function\\
    \bottomrule
  \end{tabular}
  \caption{Annotation scheme of citation functions in \citet{teufel2006automatic}.\label{tab:cit_func_label}}
\end{table}

\begin{table*}\scriptsize
  \centering
  \begin{tabular}{c|c|c|c}
  \toprule
  \textbf{Query and Ground Truth}&\textbf{Result Ranking of} \texttt{w2v}&\textbf{Result Ranking of} \texttt{d2v-cac}&\textbf{Result Ranking of} \texttt{h-d2v}\\
  \midrule
  \tabincell{m{0.2\textwidth}}{\dots We also evaluate our model by computing the machine translation BLEU score~\citep{papineni2002bleu} using the Moses system~\citep{koehn2007moses}\dots\\\\\citep{papineni2002bleu} \textbf{BLEU: a Method for Automatic Evaluation of Machine Translation}\\\citep{koehn2007moses} \textbf{Moses: Open Source Toolkit for Statistical Machine Translation}}&\tabincell{m{0.24\textwidth}}{1. HMM-Based Word Alignment in Statistical Translation\\2. Indirect-HMM-based Hypothesis Alignment for Combining Outputs from Machine Translation Systems\\3. The Alignment Template Approach to Statistical Machine Translation\\\quad\quad\quad\quad\quad\quad\quad\dots\\9. \textbf{Moses: Open Source Toolkit for Statistical Machine Translation}\\57. \textbf{BLEU: a Method for Automatic Evaluation of Machine Translation}}&\tabincell{m{0.24\textwidth}}{1. Discriminative Reranking for Machine Translation\\2. Learning Phrase-Based Head Transduction Models for Translation of Spoken Utterances\\3. Cognates Can Improve Statistical Translation Models\\\quad\quad\quad\quad\quad\quad\quad\dots\\6. \textbf{BLEU: a Method for Automatic Evaluation of Machine Translation}\\29. \textbf{Moses: Open Source Toolkit for Statistical Machine Translation}}&\tabincell{m{0.2\textwidth}}{1. \textbf{BLEU: a Method for Automatic Evaluation of Machine Translation}\\2. Statistical Phrase-Based Translation\\3. Improved Statistical Alignment Models\\4. HMM-Based Word Alignment in Statistical Translation\\5. \textbf{Moses: Open Source Toolkit for Statistical Machine Translation}}\\
  \bottomrule
  \end{tabular}
  \caption{Papers recommended by different approaches for a citation context in~\citet{zhao2010fast}.}\label{tab:case}
\end{table*}

\subsubsection{Impact of Context Intent Awareness}\label{ssec:impact_context_intent_awareness}

In this section, we analyze the impact of context intent awareness.
We use~\citet{teufel2006automatic}'s 2,824 citation contexts\footnote{The number is 2,829 in the original paper. The inconsistency may be due to different regular expressions we used.} with annotated citation functions, \eg emphasizing weakness (Weak) or using tools/algorithms (PBas) of the cited papers.
Table~\ref{tab:cit_func_label} from~\citet{teufel2006automatic} describes the full annotating scheme.
\citet{teufel2006automatic} also use manual features to evaluate citation function classification.
To test all models on capturing context intents, we average all context words' IN vectors (trained on DBLP) as features.
Noticing that \texttt{pv-dbow} does not output IN word vectors, and OUT vectors do not provide reasonable results, we use \texttt{pv-dm} here instead.
We use SVM with RBF kernels and default parameters.
Following \citet{teufel2006automatic}, we use 10-fold cross validation.

\begin{figure}
  \centering
  \includegraphics[width=0.9\columnwidth]{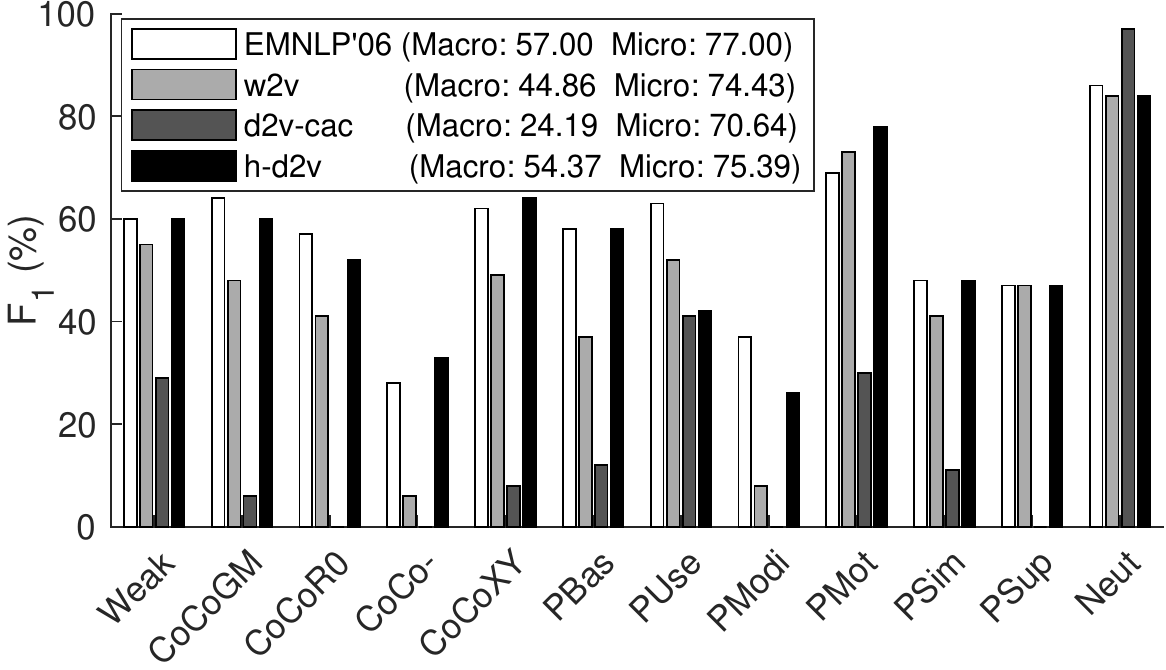}
  \caption{F\textsubscript{1} of citation function classification.}\label{fig:cit_func_clf}
\end{figure}

Figure~\ref{fig:cit_func_clf} depicts the F\textsubscript{1} scores.
Scores of~\citet{teufel2006automatic}'s approach are from the original paper.
We omit \texttt{d2v-nc} because it is very inferior to \texttt{d2v-cac}.
We have the following observations.

First, \citet{teufel2006automatic}'s feature-engineering-based approach has the best performance.
Note that we cannot obtain their original cross validation split, so the comparison may not be fair and is only for consideration in terms of numbers.

Second, among all embedding-based methods, \texttt{h-d2v} has the best citation function classification results, which is close to \citet{teufel2006automatic}'s.

Finally, the \texttt{d2v-cac} vectors are only good at Neutral, the largest class.
On the other classes and global F\textsubscript{1}, they are outperformed by \texttt{w2v} vectors.

\begin{figure}
  \centering
  \includegraphics[width=0.95\columnwidth]{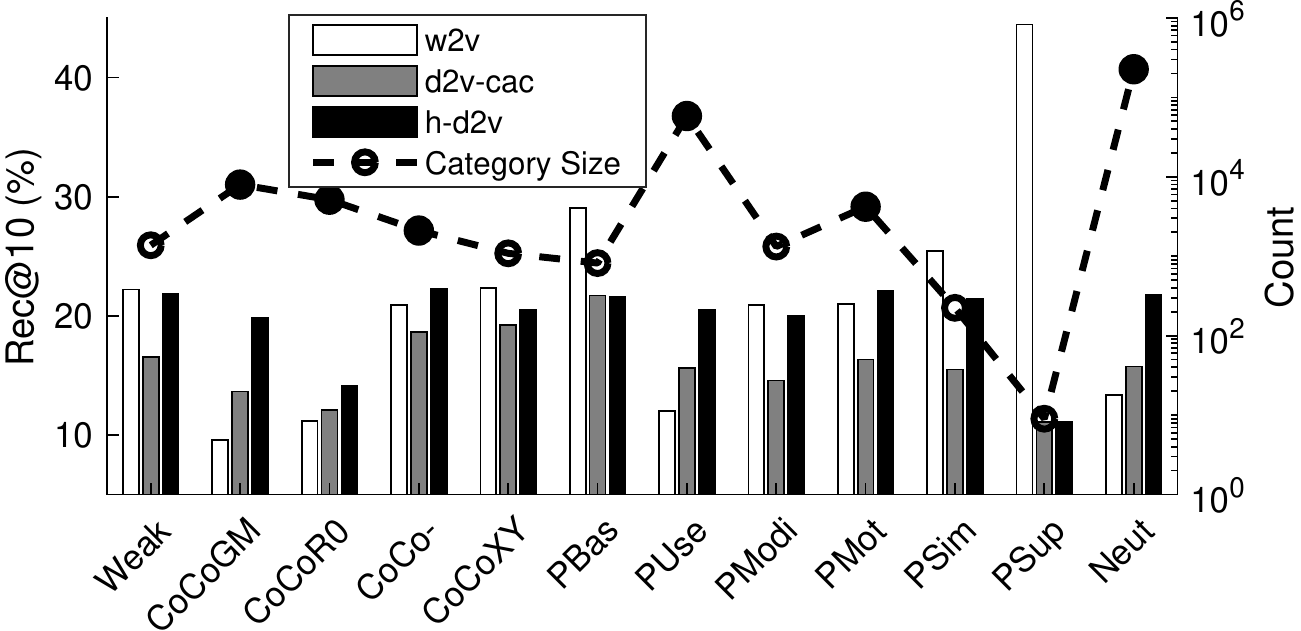}
  \caption{Rec@10 w.r.t.\ citation functions.}\label{fig:cit_recom_per_cit_func}
\end{figure}

To study how citation function affects citation recommendation, we combine the 2,824 labeled citation contexts and another 1,075 labeled contexts the authors published later to train an SVM, and apply it to the DBLP testing set to get citation functions.
We evaluate citation recommendation performance of \texttt{w2v} (I4O), \texttt{d2v-cac}, and \texttt{h-d2v} on a per-citation-function basis.
In Figure~\ref{fig:cit_recom_per_cit_func}, we break down Rec@10 scores on citation functions.
On the six largest classes (marked by solid dots), \texttt{h-d2v} outperforms all competitors.

To better investigate the impact of context intent awareness, Table~\ref{tab:case} shows recommended papers of the running example of this paper.
Here, \citet{zhao2010fast} cited the BLEU metric~\citep{papineni2002bleu} and Moses tools~\citep{koehn2007moses} of machine translation.
However, the additional words ``machine translation'' lead both \texttt{w2v} and \texttt{d2v-cac} to recommend many machine translation papers.
Only our \texttt{h-d2v} manages to recognize the citation function ``using tools/algorithms (PBas)'', and concentrates on the citation intent to return the right papers in top-5 results.

\section{Conclusion}

We focus on the hyper-doc embedding problem.
We propose that hyper-doc embedding algorithms should be content aware, context aware, newcomer friendly, and context intent aware.
To meet all four criteria, we propose a general approach, \texttt{hyperdoc2vec}, which
assigns two vectors to each hyper-doc and models citations in a straightforward manner.
In doing so, the learned embeddings satisfy all criteria, which no existing model is able to.
For evaluation, paper classification and citation recommendation are conducted on three academic paper datasets.
Results confirm the effectiveness of our approach.
Further analyses also demonstrate that possessing the four properties helps \texttt{h-d2v} outperform other models.

\bibliographystyle{acl_natbib}

\end{document}